\documentclass[10pt,twocolumn]{article}

\usepackage[utf8]{inputenc}
\usepackage[T1]{fontenc}
\usepackage{lmodern}
\usepackage{geometry}
\usepackage{amsmath, amssymb, amsfonts}
\usepackage{graphicx}
\usepackage{hyperref}
\usepackage{authblk}
\usepackage{titlesec}
\usepackage[round]{natbib}
\usepackage{rotating} 
\usepackage{placeins}
\usepackage{color,soul}
\usepackage{booktabs}
\usepackage{dblfloatfix}
\usepackage{afterpage}
\usepackage{caption}

\titleformat{\section}{\large\bfseries}{\thesection}{1em}{}
\titleformat{\subsection}{\normalsize\bfseries}{\thesubsection}{1em}{}
 
\title{\textbf{Late Fusion Neural Operators for Extrapolation Across Parameter Space in Partial Differential Equations}}

\author[1,2,*]{Eva van Tegelen}
\author[1]{Taniya Kapoor}
\author[2]{George A.K. van Voorn}
\author[2]{Peter van Heijster}
\author[1]{Ioannis N. Athanasiadis}
\affil[1]{Artificial Intelligence Group, Wageningen University and Research, Netherlands}
\affil[2]{Mathematical and Statistical Methods, Wageningen University and Research, Netherlands}
\affil[*]{\texttt{eva.vantegelen@wur.nl}}
\date{} 

\begin{document}
\twocolumn[
\maketitle

\begin{center}
{\large \bfseries Abstract}
\vspace{0.5em}

\begin{minipage}{0.82\textwidth}
\small
Developing neural operators that accurately predict the behaviour of systems governed by partial differential equations (PDEs) across unseen parameter regimes is crucial for robust generalization in scientific and engineering applications. In practical applications, variations in physical parameters induce distribution shifts between training and prediction regimes, making extrapolation a central challenge. As a result, the way parameters are incorporated into neural operator models plays a key role in their ability to generalize, particularly when state and parameter representations are entangled. In this work, we introduce the Late Fusion Neural Operator, an architecture that disentangles learning state dynamics from parameter effects, improving predictive performance both within and beyond the training distribution. Our approach combines neural operators for learning latent state representations with sparse regression to incorporate parameter information in a structured manner. Across four benchmark PDEs including advection, Burgers, and 1D/2D reaction–diffusion equations, the proposed method consistently outperforms Fourier Neural Operator and CAPE-FNO. Late Fusion Neural Operators achieve consistently the best performance in all experiments, with an average RMSE reduction of 72.9\% in-domain and 71.8\% out-domain compared to the second-best method. These results demonstrate strong generalization across both in-domain and out-domain parameter regimes.
\end{minipage}
\end{center}

\vspace{1.0em}
]

\section{Introduction}

Many physical systems can be described by partial differential equations (PDEs) that represent their spatial and temporal evolution, including applications in fluid dynamics, heat transfer, and biological and chemical processes \citep{Becker2015-lu, Markowich2016-im, Gardner2025-df}. These equations typically depend on underlying parameters representing quantities such as material properties~\citep{Evans2022-lj, pletcher2012computational, gurtin2010mechanics} and environmental conditions~\citep{vallis2017atmospheric, borgogno2009mathematical}.
In practice, parameters vary across operating regimes or environments, while data are typically available only for a limited subset of the parameter space due to physical, logistical, or economic constraints. From a machine-learning perspective, variations in physical parameters effectively change the distribution of the input data between training and prediction, a phenomenon commonly referred to as covariate shift, which poses a major challenge for data-driven models~\citep{tamang2025handling, bickel2009discriminative}. Therefore, developing models that can accurately predict system dynamics for unseen system parameter values is crucial for robust and reliable generalization in scientific and engineering applications~\citep{Li2025-lz}.

Recent advances in machine learning have introduced frameworks for learning mappings between function spaces from data, enabling data-driven solutions of PDEs \citep{Luo2026-cq,Azizzadenesheli2024-lw, Kovachki2023-zv, fanaskov2023spectral}. Neural operators, including architectures such as Fourier Neural Operator (FNO) \citep{Li2020-qz} and DeepONet \citep{Lu2019-us}, learn solution operators that can generalize across varying initial conditions and spatial resolutions, while methods such as PDE-Refiner \citep{Lippe2023-rq} incorporate iterative refinement strategies to improve long-term rollout accuracy.

A straightforward approach to apply these methods to parameter-dependent PDEs is to provide the governing parameters as additional input channels alongside the system state. In this formulation, the operator jointly processes both the evolving state and the parameter information to predict the next system update. Several recent works have proposed more specialized conditioning strategies that integrate parameter information through dedicated architectural components rather than simple input augmentation. For example, \cite{Li2025-xd} combine operator mixtures with parameter-fusion to improve out-of-distribution generalization, while \cite{Kassai-Koupai2024-dl} introduce adaptive conditioning mechanisms that adjust the solver to unseen parameter settings through short adaptation trajectories. Other approaches incorporate parameter dependence through dedicated architectural modules, such as dual-network decompositions of spatial representations and parameter-temporal dependencies \citep{Xiang2025-tp} or parameter-guided channel attention mechanisms \citep{Takamoto2023-my}.

Despite these advances, parameter information is still typically incorporated directly into a neural operator. While such approaches allow to learn complex interactions between parameter effects and state dynamics, we argue that they may hinder robust extrapolation beyond the training regime. To address this, we introduce a Late Fusion Neural Operator, inspired by classical numerical solvers~\citep{ames2014numerical}, which processes parameter information separately from the evolving state and combines them only at a later stage. By maintaining this separation, the model can learn parameter dependencies in a more structured manner, thus facilitating generalization to unseen parameter values.

In this work, we focus on in- and out-of-distribution extrapolation across the parameter space in a strictly one-shot framework, where predictions are made using only the initial system state, without access to temporal history or any form of model warm-up at inference time. When predicting across new parameter regimes, past trajectories may be limited or unavailable. While previous approaches often rely on temporal history, such as sliding windows with multiple time step inputs \citep{Li2020-qz}, or require test-time fine-tuning to adapt to new environments \citep{Kassai-Koupai2024-dl}, here we focus on predicting system dynamics solely from the initial condition and known system parameters. We furthermore assume no prior knowledge of the governing equations, including their functional form or differential operators, and model the dynamics entirely from data. 

Our main contributions can be summarized as follows:
\begin{itemize}
\item We propose a framework that couples neural operators with a late-fusion parameter module, using sparse regression to explicitly incorporate parameter dependencies in a structured and interpretable manner.

\item We benchmark the proposed method on a diverse set of PDE problems, including one- and two-dimensional systems with varying boundary conditions, and compare against standard neural operator baselines, with particular emphasis on generalization to unseen parameter regimes.

\item We perform an in-depth analysis of the model through ablation studies to quantify the contribution of individual architectural components.

\item We conduct an interpretation study of the learned operator, demonstrating how it reveals relationships between system dynamics and governing parameters, highlighting the potential of the approach for dynamic discovery.
\end{itemize}

\section{Problem set-up}

\begin{figure*}
    \centering
    \includegraphics[width=\textwidth]{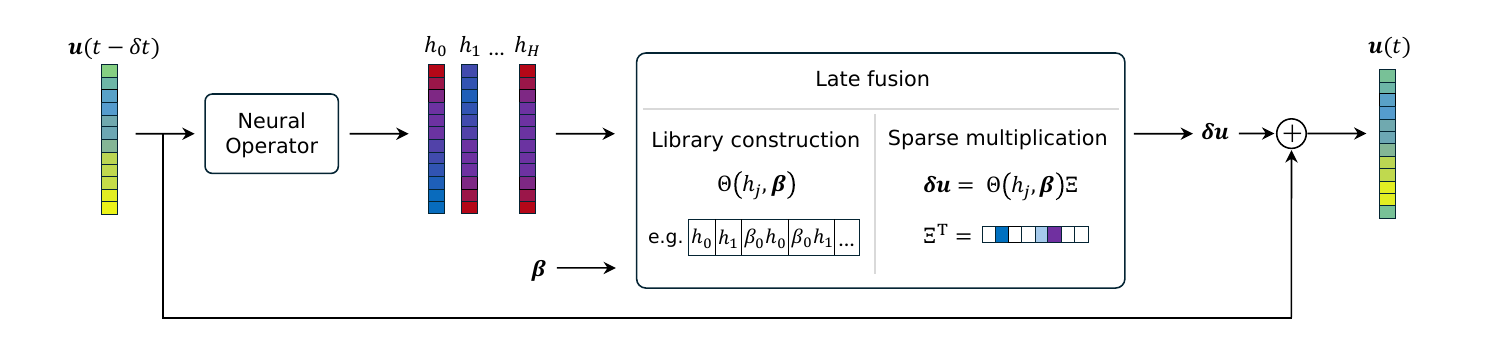}
    \caption{
    Late Fusion Neural Operator model structure. The current state $\boldsymbol{u}(t-\delta t)$ (shown here as a single system state with color indicating its value or density at each spatial location) is given as input to the neural operator, which outputs a set of hidden states $h_j$. These hidden states are combined with the input parameters $\boldsymbol{\beta}$ to construct a library vector $\Theta$. This library is then multiplied by a sparse matrix $\Xi$ to compute the rate of change $\boldsymbol{\delta u}$, which is added to the current state to obtain the next state $\boldsymbol{u}(t)$.
    }
    \label{fig:latefusion}
\end{figure*}

Following the notation of \cite{Brandstetter2022-gt}, we consider PDEs defined over a time interval $t \in [0, T]$ and over $D$ spatial dimensions $\boldsymbol{x} = \left[x_1, ..., x_D\right]^T \in \mathbb{X} \subseteq \mathbb{R}^D$, where $\mathbb{X}$ denotes the spatial domain. Furthermore, we introduce a parameter vector $\boldsymbol{\beta} \in \mathbb{P} \subseteq \mathbb{R}^P$ which characterizes the system dynamics. In our work we focus specifically on a generalized class of PDEs, of which the system state $\boldsymbol{u}: [0, T] \times \mathbb{X} \rightarrow \mathbb{R}^V$ is governed by a parameterized equation of the form:
\begin{align}
    \partial_t \boldsymbol{u}(t, \boldsymbol{x}) &= F\left(t,\boldsymbol{x}, \boldsymbol{u}, \partial_{\boldsymbol{x}} \boldsymbol{u}, \partial_{\boldsymbol{x}\boldsymbol{x}} \boldsymbol{u}, \dots; \boldsymbol{\beta} \right), \label{eq:pde_dynamics} \\
    \boldsymbol{u}(0, \boldsymbol{x}) &= \boldsymbol{u}^0(\boldsymbol{x}), \quad \boldsymbol{x} \in \mathbb{X}, \label{eq:initial_condition} \\
    B[\boldsymbol{u}](t, \boldsymbol{x}) &= 0, \quad \boldsymbol{x} \in \partial\mathbb{X}, \label{eq:boundary_condition}
\end{align}
where $F(\cdot; \boldsymbol{\beta})$ is a nonlinear differential operator parameterized by $\boldsymbol{\beta}$. Here, $\partial_t \boldsymbol{u}$ is the partial derivative of the solution $u$ with respect to time
and $\partial_{\boldsymbol{x}} \boldsymbol{u}, \partial_{\boldsymbol{x}\boldsymbol{x}}\boldsymbol{u}$,... is short for partial derivatives $\partial \boldsymbol{u}/\partial \boldsymbol{x}$,  $\partial^2 \boldsymbol{u}/\partial \boldsymbol{x}^2$ and so forth. The initial condition is denoted by $\boldsymbol{u}^0(\boldsymbol{x})$ and assumed to be sufficiently smooth. The boundary condition on the domain boundary $\partial\mathbb{X}$ is denoted by$B[\boldsymbol{u}](t, \boldsymbol{x})$  . 

We assume access to a dataset consisting of multiple ($N$) solution trajectories of the parameterized PDE corresponding to different initial conditions $\boldsymbol{u}^0(\boldsymbol{x})$ and parameter values $\boldsymbol{\beta}$, where the parameters are assumed to remain constant over time within each trajectory. Specifically, for a finite set of parameters $\{\boldsymbol{\beta}^{(i)}\}_{i=1}^{N} \subset \mathbb{P}$ and corresponding initial conditions $\{\boldsymbol{u}^{0,(i)}\}_{i=1}^{N}$, we observe $N$ trajectories
\[
\boldsymbol{u}^{(i)}(t,\boldsymbol{x}) \in \mathbb{R}^V, 
\quad t \in [0,T], \ \boldsymbol{x} \in \mathbb{X},
\]
which are discretized in time and space, yielding sequences of spatio-temporal data. 
The resulting dataset can thus be written as
\begin{align} 
\mathcal{D} = \left\{ 
\left(\boldsymbol{u}^{0,(i)}, \boldsymbol{\beta}^{(i)}, \boldsymbol{u}^{(i)}(t,\boldsymbol{x})\right) 
\right\}_{i=1}^N,
\end{align}
where each sample represents a full discretized solution trajectory conditioned on its corresponding initial condition and system parameters.

\section{Late Fusion Neural Operator}

For the Late Fusion Neural Operator, we were inspired by classical numerical solvers for PDEs, which typically decompose time integration into two conceptual steps~\citep{ames2014numerical, Gardner2025-df}. First, relevant spatial derivatives are approximated numerically from the current state of the system. Second, these numerical derivatives are combined, together with known parameters of the governing equation, to approximate the temporal derivative of the state. In the present setting however, the governing equation is assumed to be unknown, such that neither the functional form of this update rule nor the specific spatial derivatives to be approximated are available. 

To address the lack of governing equations for real world systems, we propose a data-driven framework that mirrors the two-stage structure of classical PDE solvers while remaining fully equation-agnostic. Given the current system state $\boldsymbol{u}(t-\delta t)$, where $\delta t$ denotes the temporal discretization step, we first apply a neural operator $\mathcal{N}_\theta$, parameterized by trainable weights $\theta$, to obtain a set of latent representations:
\begin{equation}
\{ h_j \}_{j=1}^H = \mathcal{N}_\theta(\boldsymbol{u}(t-\delta t)),
\end{equation}
where the hidden states $h_j$ are learned directly from data and can be viewed as analogues of intermediate quantities that arise in classical PDE discretizations, such as spatial derivatives and nonlinear interaction terms, without imposing any explicit physical interpretation.

In a second step, the late-fusion step, the hidden states are combined with known parameters values $\boldsymbol{\beta}$ via a library of candidate functions $\Theta$ and a coefficient matrix~$\Xi$ that is learned from data. This approach is inspired by the Sparse Identification of Nonlinear Dynamics (SINDy) framework~\citep{Rudy2017-jx,Brunton2016-ds}, in which governing equations are expressed as linear combinations of functions taken from a predefined library evaluated on the system state. The construction of the library of candidate functions $\Theta$ is highly flexible and can be adapted to the desired structure of the system. In practice, the library may contain polynomial terms of varying order, trigonometric functions, exponential terms, as well as interaction terms between the hidden states and the known parameters, e.g.
\begin{equation}
  \Theta(\boldsymbol{h},\boldsymbol{\beta}) =
\left\{
1,\ h_1,\ \beta_1 h_2,\ \beta_2 h_3^2,\ \beta_3 \sin(h_4 ),... 
\right\}.
\end{equation}

While the exact derivatives and terms of the system are unknown, this flexible library allows us to control the complexity of the candidate dynamics, such as the number of hidden states, the order of polynomial terms, and the form of interactions with parameters, thereby incorporating prior knowledge without requiring access to the true governing equations.

The state increment (residual) $\boldsymbol{\delta u}$ is modelled using the candidate library $\Theta(\boldsymbol{h},\boldsymbol{\beta})$, as:
\begin{equation}
\boldsymbol{\delta  u} = \Theta(\boldsymbol{h},\boldsymbol{\beta})\,\Xi,
\end{equation}
where $\Xi$ is a trainable coefficient matrix whose entries weigh the contributions of the candidate functions in  the library $\Theta(\boldsymbol{h},\boldsymbol{\beta})$. Sparsity is encouraged through regularization during training, allowing the model to select a subset of candidate functions without imposing hard thresholding.
The state is then advanced in time according to
\begin{equation}
\boldsymbol{u}(t) = \boldsymbol{u}(t-\delta t) +  \boldsymbol{\delta u}.
\end{equation}

The model is trained using a composite loss function that balances predictive accuracy with sparsity of the late-fusion module. The first component of the loss calculates the Mean Squared Error (MSE) between the predicted state and the observed data, and is defined as
\begin{equation}
\mathcal{L}_{\mathrm{data}}
= \frac{1}{N}\sum_{i=1}^{N}
\left\lVert 
\boldsymbol{u}_{\mathrm{pred}}^{(i)}
-
\boldsymbol{u}^{(i)}_{\mathrm{true}}
\right\rVert_{2}^{2},
\end{equation}
where $\boldsymbol{u}^{(i)}_{\mathrm{pred}}$ denotes the predicted state for sample $i$ and $\boldsymbol{u}^{(i)}_{\mathrm{true}}$ the corresponding ground-truth observation. A second loss term enforces sparsity on the coefficient matrix $\Xi$. Specifically, we employ $\ell_1$-regularization (Lasso \citep{Hastie2003-oc}),
\begin{equation}
\mathcal{L}_{\mathrm{sparse}} = \lVert \Xi \rVert_1,
\end{equation}
with the aim of only a small subset of candidate functions from the library. The total training objective is then given by:
\begin{equation}
\mathcal{L} = \mathcal{L}_{\mathrm{data}} + \lambda_{\mathrm{sparse}}\,\mathcal{L}_{\mathrm{sparse}}, \label{eq:loss}
\end{equation}
where the sparsity coefficient $\lambda_{\mathrm{sparse}}$ controls the trade-off between predictive accuracy and model simplicity. This coefficient is selected via hyperparameter tuning in a valiation split(see Appendix~\ref{app:hyperparameters}).

It is important to note that our approach shares the high-level idea of late parameter incorporation with recent work such as iMOEE \citep{Li2025-xd}, which also separates state dynamics from parameter effects. However, iMOEE differs from our method in multiple aspects. First, it employs a mixture-of-experts operator architecture, whereas we use a single neural operator backbone. Second, parameter information is incorporated in iMOEE via an MLP-based fusion mechanism, while we instead use a structured sparse regression library for a more explicit and interpretable parameter representation. Finally, their approach introduces additional components, including precomputed derivative features and invariance-based training objectives, whereas our framework does not rely on such additional inputs or training objectives.

\section{Experiments}

To validate the proposed approach and evaluate its in- and out-of-distribution prediction capabilities, we generate synthetic datasets from a subset of representative PDEs in one and two dimensions taken from the benchmark framework PDEBench \citep{Takamoto2022-nl}: the 1D advection equation, the 1D Burgers equation, the 1D reaction-diffusion equation, and the 2D reaction-diffusion equation. This subset covers different dynamical behaviours, such as shocks and spatial pattern formation, and provides a balanced trade-off between diversity and computational manageability. We use the PDEBench framework to generate all datasets; details on the numerical schemes can be found in the original paper. As the original PDEBench datasets are not designed for systematic evaluation across a wide range of parameter values, we adapted the original generation scripts to sample for a wider range of parameter values and create separate training and test distributions. 

For all datasets we sample system parameter values from a uniform distribution. Table~\ref{tab:libraries_and_ranges} describes the ranges of the parameter values that were used to generate the training and test datasets. Since our model is trained on single-step predictions and we do not employ specialized stabilization techniques, we restrict the parameter ranges to regimes that yield numerically stable solutions and avoid error blow-up. Visualisations of the data can be found in Appendix~\ref{app:visuals}.

\begin{table*}[t!]
\centering
\caption{Libraries together with the parameter ranges for in-domain and out-domain sampling. Parameters are sampled from a uniform distribution with the specified ranges. For training and in-domain testing, 100 parameter values are sampled from the in-domain range. Out-domain testing uses samples from the out-domain range.}
\vspace{0.5em}
\begin{tabular}{lllll}
\toprule
\textbf{Equation} & \textbf{Parameter} & \textbf{In-domain} & \textbf{Out-domain} & \textbf{Library} \\
\midrule

1D advection 
& $\beta$ 
& $(0, 0.5)$ 
& $(0.5, 1)$
& $\{h_0\beta,\ h_1\}$ \\

1D Burgers 
& $\nu$ 
& $(0.01, 0.02)$ 
& $(0, 0.01)$
& $\{h_0\nu,\ h_1\}$ \\

1D reaction-diffusion 
& $\rho$ 
& $(0, 1)$ 
& $(0, 1)$
& $\{1,\ h_0,\ h_1,\ h_0^2,\ h_1^2,\ h_0h_1,$ \\

& $\nu$ 
& $(0, 0.1)$ 
& $(0.1, 0.2)$
& $\rho h_0^2,\ \rho h_1^2,\ \rho h_0h_1, \ \nu h_0^2,\ \nu h_1^2,\ \nu h_0h_1\}$ \\

2D reaction-diffusion 
& $k$ 
& $(0, 0.05)$ 
& $(0.05, 0.075)$
& $\{1,\ h_0,\ h_1, \ h_2, \ k,\ kh_0,\ kh_1,\ kh_2\}$ \\

\bottomrule
\end{tabular}
\label{tab:libraries_and_ranges}
\end{table*}

\subsubsection*{1D advection equation}
We consider the one-dimensional advection equation:
\begin{align}
    \partial_t u(t,x) &=-\beta \partial_x u(t,x) ,\ x \in (0,1), t\in(0,0.5],\\
    u(0,x)&=u_0(x), \quad x\in(0,1),
\end{align}
where $\beta$ is a constant advection speed. We use the periodic boundary conditions and initial conditions as implemented in \cite{Takamoto2022-nl}:
\begin{equation}
u_0(x) = \sum_{i=1}^{N} A_i \sin\left(k_i x + \phi_i\right),\label{eq:initialconditions}
\end{equation}
where $k_i =2\pi\{n_i\}/L_x$ are the wave numbers with $\{n_i\}$ as integer values selected randomly from the interval $[1, n_{\max}]$, $N$ is the number of waves to be added, and $L_x$ is the length of the spatial domain. The amplitudes $A_i$ are sampled from a uniform distribution $(0,1)$, and the phases $\phi_i$ are random values in $(0,2\pi)$. We set $k_{\max} = 8$ and $N = 2$.

We generate data for different values of the parameter $\beta$. For training and in-domain extrapolation, we randomly sample $\beta$ from the uniform distribution with interval $(0, 0.5)$. Out-domain data are generated by sampling $\beta$ from the interval $(0.5, 1)$. We select training and test intervals to assess out-of-distribution performance: the test set includes higher advection speeds than seen during training. All trajectories are sampled over the time interval $[0, 0.5]$ with a fixed time step of $\Delta t = 0.05$. We use a spatial resolution of 128 grid points.

\subsubsection*{1D Burgers equation} 
We consider the one-dimensional Burgers equation:
\begin{align}
    \partial_t u(t,x) &=-\partial_x( u^2(t,x)/2) +\nu\pi \partial_{xx}u(t,x) , \\
      & \quad\quad\quad\quad\quad\quad\quad x \in (0,1), t\in(0,0.5],\nonumber\\
    u(0,x)&=u_0(x), \quad x\in(0,1),
\end{align}
where $\nu$ is the diffusion coefficient, which we vary. We use periodic boundary conditions and initial conditions (Eq.~\ref{eq:initialconditions}) following the Burgers setup in \cite{Takamoto2022-nl}. For training and in-domain extrapolation, we randomly sample $\nu$ from the interval $(0.01,0.02)$. Out-domain data are generated by sampling $\nu$ from the interval $(0,0.01)$, corresponding to weaker diffusion and therefore sharper shock formation. All trajectories are sampled over the time interval $[0, 0.5]$ with a fixed time step of $\Delta t = 0.005$ and a spatial resolution of 128.

\subsubsection*{1D reaction-diffusion equation}
Here, we consider a one-dimensional scalar reaction-diffusion equation, that combines a diffusion process and a rapid evolution from a source term, better known as the Fisher-KPP equation \citep{Fisher1937-vq}:
\begin{align}
    \partial_t u(t,x) &= \nu\partial_{xx}u(t,x) +\rho u(1-u), \\
      & \quad\quad\quad\quad\quad\quad\quad x \in (0,1), t\in(0,0.5],\nonumber\\
    u(0,x)&=u_0(x), \quad x\in(0,1),
\end{align}
where $\nu$ is the diffusion coefficient, $\rho$ the force term coefficient. We use periodic boundary conditions and initial conditions (Eq.~\ref{eq:initialconditions}) following the setup in \cite{Takamoto2022-nl}. For training and in-domain extrapolation, we randomly sample both $\nu$ and $\rho$ from the intervals $(0,0.1)$ and $(0,1)$ respectively. Out-of-domain data are generated by extrapolating in the direction of $\nu$. We sample $\nu$ from the interval $(0.1,0.2)$, but sample the force term coefficient $\rho$ from the same region as was used for training.

\subsubsection*{2D reaction-diffusion equation}
In addition to the 1D reaction-diffusion equation, we also consider a two-dimensional extension with two nonlinearly coupled variables, with Fitzhugh-Nagumo reaction functions \citep{Fitzhugh1961-aa, Klaasen1984-bi}:
\begin{align*}
    \partial_t u&=D_u \partial_{xx} u + D_u \partial_{yy} u + u -u^3-k-v,\\ 
    \partial_t v&=D_v \partial_{xx} v + D_v \partial_{yy} v + u-v,\\
    & \quad\quad x\in (-1,1), y\in (-1,1), t\in(0,4],\nonumber
\end{align*}
where the diffusion coefficient are $D_u = 1 \times  10^{-3}$ and $D_v = 5 \times  10^{-3}$ and the parameter $k$ is sampled from $[0,0.5]$ during training and in-domain testing and sampled from $[0.5,0.75]$ for out-domain testing.  Instead of using periodic boundary as was done in all other test-cases, we use no-flow Neumann boundary conditions. Initial conditions follow the setup described in \cite{Takamoto2022-nl}. The 2D reaction-diffusion equation is a challenging test-case as we are dealing with two nonlinearly coupled variables of interest, an activator and inhibitor, as well as non-periodic boundary conditions.

\medskip
For each test case, we construct candidate libraries with varying levels of complexity, as summarized in Table~\ref{tab:libraries_and_ranges}. The design of these libraries is guided both by the complexity of the underlying governing equations and by the aim of evaluating the performance of the Late Fusion Neural Operator framework under different levels of library complexity. For the 1D advection and Burgers equations, we employ relatively simple libraries consisting of a single parameter-dependent term and a single non-parameter-dependent term, reflecting the limited number of terms present in the governing equations.  In contrast, for the 1D reaction-diffusion equation, we constructed a more complex library by creating a polynomial representation of both the hidden states and the system parameters, resulting in a relatively large set of candidate terms. This was done to demonstrate that even with more complex libraries, the late-fusion framework is capable of correctly identifying the governing dynamics. For the 2D reaction–diffusion equation, we again use a polynomial library, but restrict the hidden-state representation to first-order terms to balance library complexity with the increased complexity arising from the higher-dimensional system. 

Finally, we note that the impact of library complexity is closely tied to the sparsity regularization strength: more expressive libraries remain effective provided that appropriate sparsity constraints are imposed. This relationship is further analysed in Sec.~\ref{app:libraries}, where we show how performance varies with library size and the sparsity coefficient.

We benchmark our method against two established neural operator approaches. As a standard baseline, we use the Fourier Neural Operator (FNO) \citep{Li2020-qz}, where the parameter is added as an additional input channel alongside the state variables.  We also compare against a method that was explicitly designed for parameter incorporation: CAPE \citep{Takamoto2023-my}, which uses a channel-guided attention mechanism to inject PDE parameter information into existing neural operators and solvers. All methods in this comparison use the same Fourier Neural Operator backbone to ensure a controlled and fair evaluation of parameter incorporation strategies. Additionally, we evaluate the proposed late-fusion module with an alternative neural operator backbone (CNO) in Sec.~\ref{app:cno}.

We deliberately do not benchmark against other parameter-aware PDE learning methods that differ substantially in problem formulation or introduce additional modelling components beyond parameter-fusion. In particular, methods based on online adaptation trajectories \citep{Kassai-Koupai2024-dl} do not align with our strict one-shot prediction setting. Likewise, the approach of \citet{Xiang2025-tp} combines parameter conditioning with a physics-informed framework and does not employ a neural operator backbone, making it fundamentally different from our setup. Similarly, the recent iMOOE framework \citep{Li2025-xd} combines parameter-fusion with a mixture-of-operator-experts architecture, precomputed derivative features, and specialized invariance-based training objectives, introducing multiple additional inductive biases. Including such methods would make it difficult to attribute performance differences specifically to the parameter incorporation mechanism rather than to broader architectural or training-design changes.

To ensure a fair comparison and focus specifically on how parameter information is incorporated, all methods follow the same training setup. In particular, to strictly adhere to the one-shot prediction goal, all models are trained on single-step transitions ($t \to t+\delta t$) and applied auto-regressively during inference to generate the full trajectory from the initial condition. We furthermore do not use additional training strategies such as the curriculum learning procedure proposed in \cite{Takamoto2023-my}. Further details on the training procedure and hyperparameter tuning are provided in Appendix~\ref{app:hyperparameters}.

\section{Results}

\begin{figure*}[p]
    \centering
    \includegraphics[width=1\textwidth]{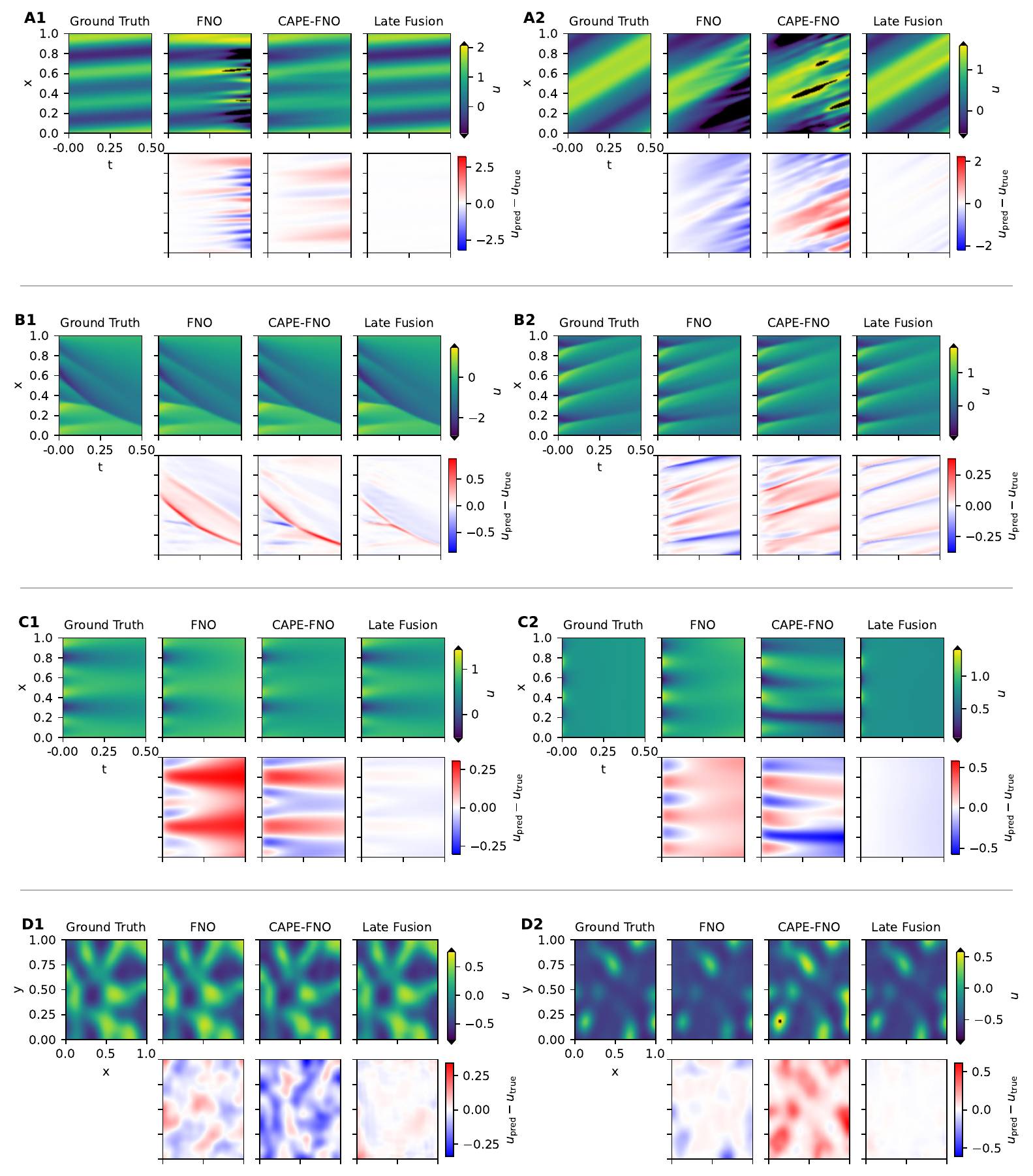}
    \caption{
    Visualisation of prediction results for  
    (A) 1D advection equation, 
    (B) 1D Burgers equation, 
    (C) 1D reaction-diffusion equation, and 
    (D) 2D reaction-diffusion equation. 
    For each equation, the left subpanel shows an in-domain prediction and the right subpanel an out-of-domain prediction. More specifically the following parameter values correspond to panels; A1: $\beta=0.11$, A2: $\beta=0.97$; 
    B1: $\nu=1.6\times10^{-2}$, B2: $\nu=9.4\times10^{-3}$; 
    C1: $\nu=3.8\times10^{-1}, \rho=1.6\times10^{-2}$, 
    C2: $\nu=5.5\times10^{-1}, \rho=1.3\times10^{-1}$; 
    D1: $k=4.1\times10^{-3}$, D2: $k=6.8\times10^{-2}$. 
    For all cases, the top row shows predicted trajectories (or the predicted solution at $T=4$ for the 2D reaction-diffusion case) for different model architectures, while the bottom row shows the corresponding error maps $u_\mathrm{pred}-u_\mathrm{true}$.
    }
    \label{fig:heatmaps}
\end{figure*}

\begin{figure*}[p]
    \centering
    \includegraphics[width=1\textwidth]{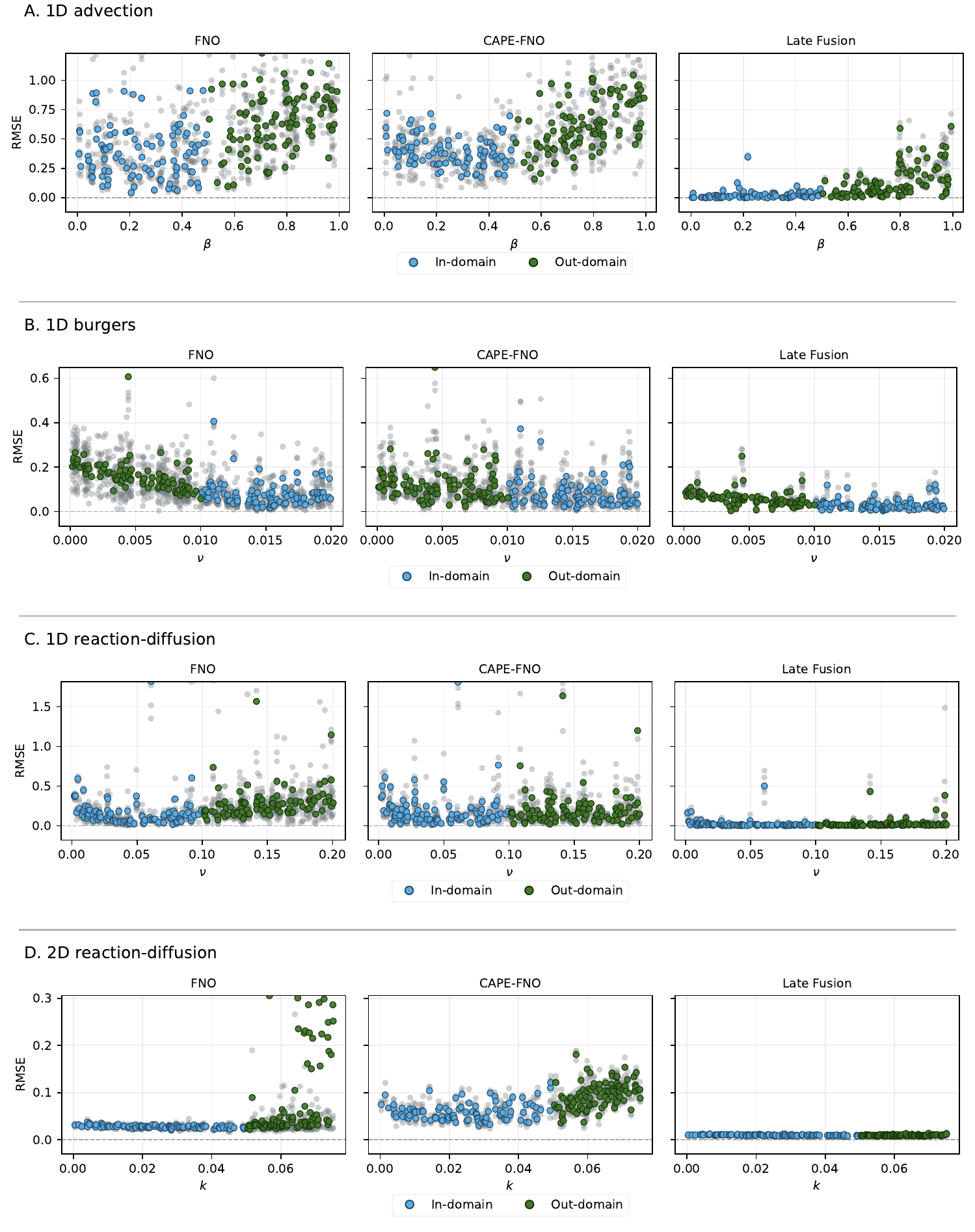}
    \caption{RMSE per trajectory for (A) 1D advection equation, 
    (B) 1D Burgers equation, 
    (C) 1D reaction-diffusion equation, and 
    (D) 2D reaction-diffusion equation. 
    RMSE per trajectory as a function of the parameter shown separately for each method. Each point corresponds to a single trajectory, with grey markers indicating results from different random seeds. The blue and green markers denote the mean RMSE across seeds for the in-domain and out-domain test sets, respectively.}
    \label{fig:results_params}
\end{figure*}

\subsection{Quantitative and qualitative performance}
We evaluate all methods on the 1D advection, 1D Burgers, 1D reaction-diffusion, and 2D reaction-diffusion test datasets using the root mean squared error (RMSE) as the primary performance metric. All reported results are averaged over five random seeds to account for training variability, with standard deviations included to indicate robustness. Additional evaluation metrics and corresponding results are provided in Table \ref{tab:extrametrics} in Appendix~\ref{appendix:extra_metrics}.

Table~\ref{tab:rmse_all} summarizes the results for both in-domain and out-of-domain evaluations. The in-domain setting measures predictive accuracy under conditions similar to those encountered during training, whereas the out-of-domain setting assesses each model’s ability to generalize to unseen parameter regimes. 
Across all equation benchmarks and for both evaluation settings, the Late Fusion Operator consistently achieves the lowest RMSE. When comparing to the second-best baseline in each setting, this corresponds to an average RMSE reduction of 72.9\% in-domain and 71.8\% out-domain, highlighting the strong accuracy and generalization performance of the proposed approach.

\begin{table}[t!]
\centering
\caption{RMSE comparison for all test cases evaluated on in-domain and out-domain settings. Lower is better and best performing model is indicated in bold. Results are reported as mean $\pm$ standard deviation.}
\vspace{0.5em}
\small
\label{tab:rmse_all}
\begin{tabular}{lcc}
\toprule
Model & In-domain & Out-domain \\
\midrule
\multicolumn{3}{l}{\textbf{1D advection}} \\
\midrule
FNO & 4.72e-1 $\pm$ 2.34e-2 & 7.14e-1 $\pm$ 1.09e-1 \\
CAPE-FNO & 4.26e-1 $\pm$ 7.69e-2 & 6.67e-1 $\pm$ 1.21e-1 \\
Late Fusion & \textbf{4.75e-2 $\pm$ 2.50e-3} & \textbf{1.93e-1 $\pm$ 2.23e-2} \\
\midrule
\multicolumn{3}{l}{\textbf{1D Burgers}} \\
\midrule
FNO & 1.02e-1 $\pm$ 3.18e-2 & 1.83e-1 $\pm$ 5.95e-2 \\
CAPE-FNO & 1.09e-1 $\pm$ 2.08e-2 & 1.50e-1 $\pm$ 4.28e-2 \\
Late Fusion & \textbf{3.91e-2 $\pm$ 2.02e-3} & \textbf{6.87e-2 $\pm$ 7.40e-4} \\
\midrule
\multicolumn{3}{l}{\textbf{1D reaction-diffusion}} \\
\midrule
FNO & 2.77e-1 $\pm$ 3.27e-2 & 3.90e-1 $\pm$ 1.30e-1 \\
CAPE-FNO & 3.01e-1 $\pm$ 1.49e-2 & 3.19e-1 $\pm$ 5.01e-2 \\
Late Fusion & \textbf{6.29e-2 $\pm$ 1.10e-2} & \textbf{8.98e-2 $\pm$ 7.10e-2} \\
\midrule
\multicolumn{3}{l}{\textbf{2D reaction-diffusion}} \\
\midrule
FNO & 2.83e-2 $\pm$ 8.23e-4 & 1.19e-1 $\pm$ 1.52e-1 \\
CAPE-FNO & 6.29e-2 $\pm$ 3.51e-3 & 9.87e-2 $\pm$ 8.43e-3 \\
Late Fusion & \textbf{1.03e-2 $\pm$ 5.86e-4} & \textbf{9.92e-3 $\pm$ 4.89e-4} \\
\bottomrule 
\end{tabular}
\end{table}

To qualitatively assess model behaviour, Figure~\ref{fig:heatmaps} compares predicted trajectories for all equations in both the in-domain and out-domain setting, together with the corresponding error maps. Across all test cases, the Late Fusion Neural Operator consistently produces predictions that more closely match the ground-truth spatio-temporal evolution than for both FNO and CAPE-FNO. This is further confirmed by the error visualisations, where the late-fusion model exhibits lower and more spatially localized errors throughout the trajectories.

Examining the individual equations reveals distinct qualitative differences between the models. For the 1D advection equation, both FNO and CAPE-FNO show progressively increasing error over time, indicating stronger error accumulation during rollout prediction. While CAPE-FNO remains relatively accurate in the in-domain setting, its performance degrades substantially in the out-of-domain regime. In contrast, the Late Fusion Neural Operator remains considerably more consistent across both settings. For the 1D Burgers equation, the largest differences appear near the shock regions, where the solution contains sharp gradients. Here, the late-fusion model produces noticeably lower local error, indicating that it is better able to preserve the sharpness of the evolving shock fronts. For the 1D reaction–diffusion system, the out-of-domain setting exhibits faster temporal dynamics due to stronger diffusion effects. Among the compared methods, only our method is able to accurately capture this accelerated behaviour, whereas the baseline models increasingly lag behind the true solution evolution. In the 2D reaction–diffusion case, the predicted trajectories of all methods appear visually similar at first glance. However, the differences in model performance become much clearer in the corresponding error maps, which reveal substantially lower errors for the Late Fusion Neural Operator.

\subsection{Performance across parameter space}
Figure \ref{fig:results_params} reports the RMSE per trajectory on the $y$-axis as a function of the corresponding physical parameter values on the $x$-axis for all equation benchmarks. Results are averaged over multiple random seeds, with gray dots indicating variability across different seeds.
By visualizing the error across the full parameter range, including both in-domain and out-of-domain regimes, Figure \ref{fig:results_params} provides a detailed view of how model performance changes as the underlying dynamics become increasingly different from those seen during training. 

\begin{table*}[b]
\centering
\caption{Libraries used to study influence of library complexity and sparsity regularization.}
\vspace{0.3em}
\small
\begin{tabular}{rllll}
\toprule
\textbf{} & \textbf{Number} & \textbf{Order } & \textbf{Order } & \textbf{Library terms} \\
\textbf{} & \textbf{of terms} & \textbf{states} & \textbf{ parameters} & \textbf{} \\
\midrule
Library 1 
& 6 & 1 & 1 & $\{1,\ h_0,\ h_1,\ \beta,\ \beta h_0,\ \beta h_1\}$ \\
Library 2 
& 9 & 1 & 2 & $\{1,\ h_0,\ h_1,\ \beta,\ \beta h_0,\ \beta h_1,\ \beta^2,\ \beta^2 h_0,\ \beta^2 h_1\}$ \\
Library 3 
& 12 & 2 & 1 & $\{1,\ h_0,\ h_1,\ h_0^2,\ h_1^2,\ h_0h_1,\ \beta,\ \beta h_0,\ \beta h_1,\ \beta h_0^2,\ \beta h_1^2,\ \beta h_0h_1\}$ \\
Library 4 
& 18 & 2 & 2 & $\{1,\ h_0,\ h_1,\ h_0^2,\ h_1^2,\ h_0h_1,\ \beta,\ \beta h_0,\ \beta h_1,\ \beta h_0^2,\ \beta h_1^2,\ \beta h_0h_1,$\\
&&& &$\beta^2,\ \beta^2 h_0,\ \beta^2 h_1,\ \beta^2 h_0^2,\ \beta^2 h_1^2,\ \beta^2 h_0h_1\}$ \\
\bottomrule
\end{tabular}\label{tab:library_complexity}
\end{table*} 

\begin{figure*}[b]
    \centering
    \includegraphics[width=1.\textwidth]{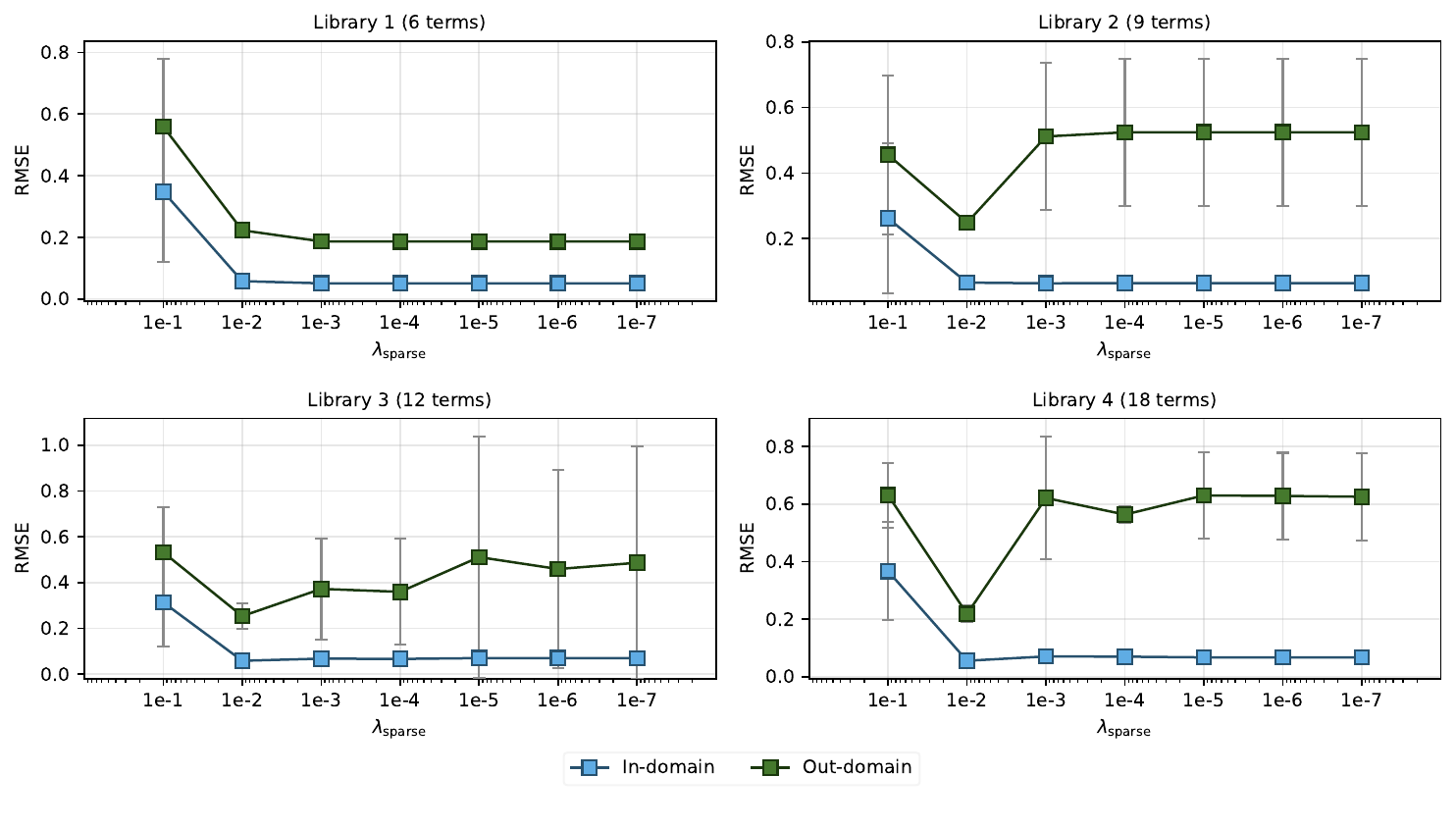}
    \caption{RMSE as a function of the sparsity coefficient $\lambda_{\mathrm{sparse}}$ for four polynomial libraries of increasing complexity, containing 6, 9, 12, and 18 candidate terms. Standard deviations are indicated by grey error bars.}
    \label{fig:fig_complexity}
\end{figure*}

Across all benchmarks, the Late Fusion Neural Operator consistently maintains lower error throughout the parameter range compared to both FNO and CAPE-FNO. In addition, the variability across random seeds remains comparatively small, indicating improved robustness to initialization and more stable performance across different training runs.

A more detailed, equation-specific analysis further highlights these trends. For the advection equation, the late-fusion model consistently achieves lower errors across the full parameter range, including both in-domain and out-of-domain regimes. While a gradual increase in error is observed in the out-of-domain region, the performance remains notably more stable compared to the baseline models. A similar behaviour is observed for the 1D Burgers equation, where the late-fusion framework again maintains lower and more consistent errors, particularly in the parameter extrapolation regime. For the 1D reaction–diffusion equation, a notable improvement of the late-fusion model is reflected in its reduced variability across seeds, alongside generally lower error levels throughout the parameter space. In the challenging 2D reaction–diffusion setting, the FNO baseline exhibits relatively stable performance within the training domain, but its error increases substantially under out-of-domain conditions, whereas the late-fusion framework again demonstrates more consistent performance and robustness.

\subsection{Influence of Library Complexity and Sparsity Regularization}\label{app:libraries}

\begin{figure*}[b]
    \centering
    \includegraphics[width=1\textwidth]{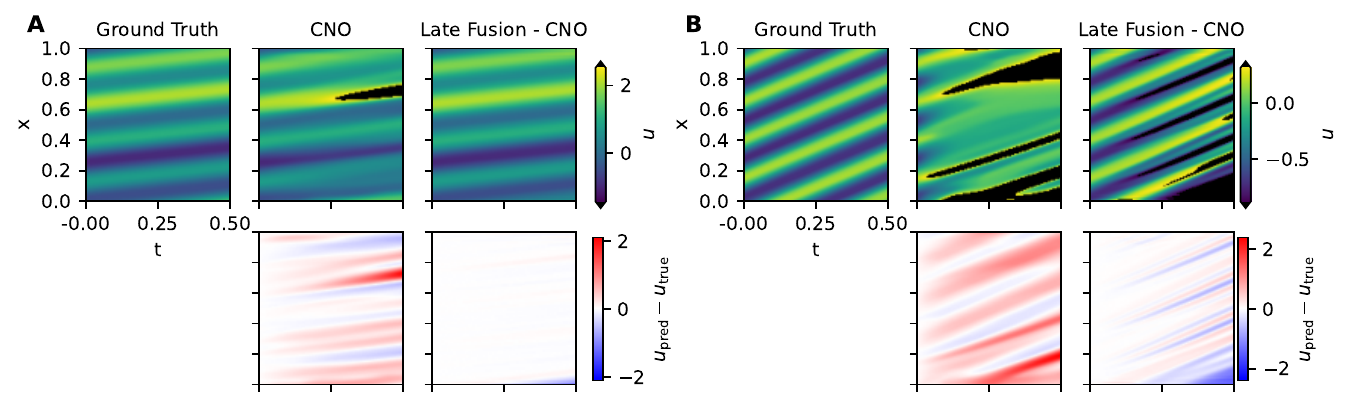}
    \caption{
    Visualisation of prediction results for  
    1D advection equation for CNO and Late-fusion CNO. The left subpanel shows an in-domain prediction ($\beta=0.11$) and the right subpanel an out-of-domain prediction ($\beta=0.97$).
    The top row shows predicted trajectories, while the bottom row shows the corresponding error maps $u_\mathrm{pred}-u_\mathrm{true}$.
    }
    \label{fig:heatmaps_CNO}
\end{figure*}

In this section, we investigate the influence of the candidate library complexity and the sparsity regularization coefficient on extrapolation performance. To this end, we consider a family of polynomial libraries in which we vary both the order of the state variables and the parameter. We consider four libraries of increasing complexity summarized in Table~\ref{tab:library_complexity}. In this instance, with increased complexity we refer to the use of higher-order polynomial terms and consequently more terms in the candidate library.

Figure~\ref{fig:fig_complexity} shows the RMSE as a function of the sparsity coefficient $\lambda_{\mathrm{sparse}}$ for all four libraries. We observe that the smallest library, containing only 6 terms, yields the best overall out-of-domain performance. In addition, its performance remains relatively stable across the full range of sparsity coefficients. This limited sensitivity is likely due to the already compact nature of the library, which restricts the number of candidate terms and therefore requires less strong regularization to generalize well.

\begin{table}[b]
\centering
\caption{RMSE comparison for advection evaluated on in-domain and out-domain settings for CNO and Late Fusion Neural Operator with CNO backbone. Lower is better and best performing model is indicated in bold. Results are reported as mean $\pm$ standard deviation.}
\label{tab:rmse_cno}
\vspace{0.5em}
\small
\begin{tabular}{lcc}

\toprule
Model & In-domain & Out-domain \\
\midrule
\multicolumn{3}{l}{\textbf{1D Advection (CNO Study)}} \\
\midrule
CNO & 5.35e-1 $\pm$ 2.30e-1 & 8.17e-1 $\pm$ 1.77e-1 \\
Late fusion- & \textbf{1.03e-1 $\pm$ 1.16e-2} & \textbf{3.47e-1 $\pm$ 3.83e-2} \\
CNO\\
\bottomrule
\label{tab:cno_results}
\end{tabular}
\end{table}
For the larger libraries, a clear dependence on the sparsity coefficient is visible. While the in-domain performance remains relatively stable across different values of the sparsity coefficient, the out-domain RMSE shows more noticeable variation, displaying the existence of an optimal sparsity coefficient for extrapolation. This highlights that, for more complex libraries, the sparsity penalty plays a crucial role in selecting only the terms that meaningfully contribute to the dynamics, while suppressing spurious terms that may still fit the training data but harm extrapolation.

These results demonstrate that larger and more complex libraries can still yield strong performance, provided that the sparsity regularization is chosen carefully. As the number of candidate terms increases, the sparsity coefficient becomes increasingly important for robust extrapolation. In addition, we observe that the standard deviation across runs tends to increase with library size, indicating that the results become less consistent and therefore less robust for larger libraries.

\subsection{Other backbone architecture: CNO}\label{app:cno}

In previous experiments, we exclusively used FNO as the backbone architecture to enable a controlled comparison and to isolate the effect of the parameter-injection strategy. This design choice allowed us to focus on how the system parameter is incorporated into the neural operator. To demonstrate that the proposed late-fusion parameterisation is not limited to FNOs, we additionally evaluate the approach using a Convolutional Neural Operator (CNO) \citep{Raoni-c2023-hl}. The goal is to show that the late-fusion mechanism can also improve performance in other neural operator architectures.

Figure~\ref{fig:heatmaps_CNO} presents a qualitative comparison between CNO, in which the parameter is introduced as an additional input channel, and the late-fusion CNO variant. Although the overall prediction quality remains below that of the FNO results shown in Figure~\ref{fig:heatmaps}, the visualizations clearly indicate that the late-fusion approach substantially improves the prediction accuracy compared to the naive parameter-injection strategy. This improvement is further confirmed quantitatively in Table~\ref{tab:cno_results}, where the late-fusion CNO achieves lower RMSE values than the classic CNO in both the in-domain and out-domain settings. These results show that the benefits of the late-fusion framework generalize beyond FNO and extend to other neural operator families.

\begin{figure*}[p]
    \centering
    \includegraphics[width=1\textwidth]{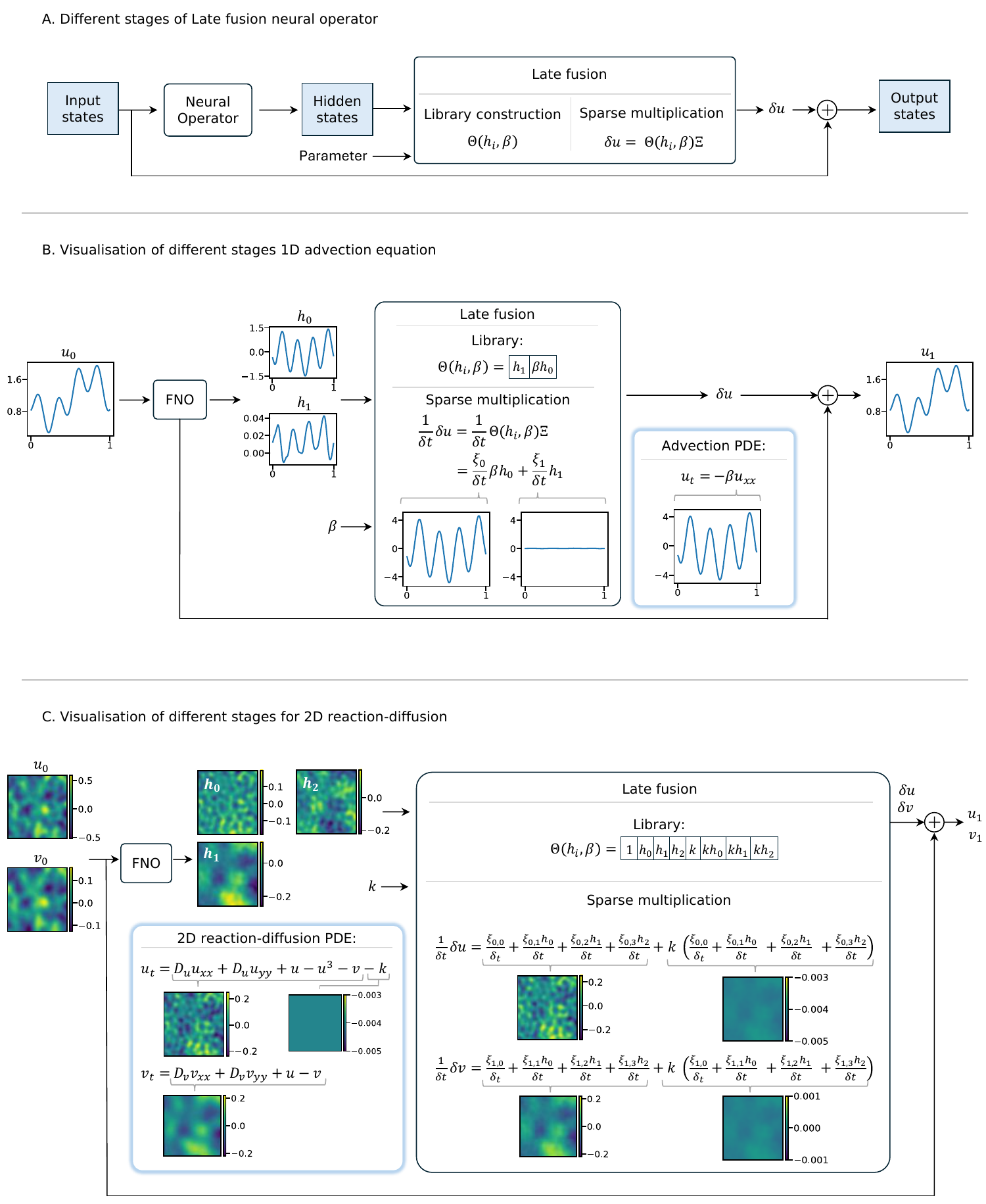}
    \caption{(A) Schematic overview of the Late Fusion Neural Operator. The input field $u_0$ is processed by the FNO, producing intermediate hidden states that are used to construct a feature library. This library is linearly combined via a learned coefficient matrix to produce the residual.
    (B–C) Visualisations of the hidden states generated by the network for a given input $u_0$ for 1D advection equation and 2D reaction-diffusion equation. The blue boxes show finite-difference approximations of derivatives of $u_0$ from the underlying PDE; these are not part of the model pipeline, but are included to highlight the correspondence between the learned features and the analytical structure of the original equation.}
    \label{fig:interprate_adv}
\end{figure*}

\newpage
\subsection{Interpretability analysis}

We perform interpretability analysis for the 1D advection and 2D reaction–diffusion equations to illustrate how different learned systems can be interpreted through their learned sparse representations. We use the same structure introduced in Figure \ref{fig:latefusion} to visualize the outputs of the different components of the late-fusion model in Figure \ref{fig:interprate_adv}. 

For the 1D advection equation, starting from an initial condition $u_0$, we propagate the input through the model to obtain the corresponding hidden representations. These hidden states are then used to construct a predefined candidate library consisting of two terms: one parameter-dependent term and one parameter-independent term, allowing us to separately analyse their contributions to the inferred dynamics.

When visualizing the parameter-dependent component of the resulting sparse regression, we observe that it closely matches the structure of the advection operator, effectively resembling the discretized derivative term associated with the underlying PDE for the given $u_0$. In contrast, the parameter-independent component remains approximately zero across the domain, which is consistent with the absence of non-parameter-driven dynamics in the advection equation. Notably, despite the model being fully data-driven, the recovered structure clearly reflects the underlying advection PDE. Overall, this indicates that the learned equation is primarily driven by a parameter-dependent term consistent with the advection operator, suggesting that the model extracts interpretable information about how the dynamics depend on the underlying parameter.

For the 2D reaction–diffusion system, we perform a similar interpretability analysis using the learned late-fusion operator. In this case, the model takes two initial states, $u_0$ and $v_0$, and produces three hidden representations through the FNO backbone. Using these hidden states together with the physical parameter $k$, we construct a candidate library which is then combined with a learned coefficient matrix to form the resulting residual equations for both $u$ and $v$. Each residual equation can be separated into a parameter-dependent and a parameter-independent component, allowing us to analyse their respective roles in the inferred dynamics.
\FloatBarrier
For the $u$-component, the parameter-independent term aligns with the non-parameter-driven part of the original reaction–diffusion PDE, while the parameter-dependent term closely matches the expected diffusion contribution proportional to $-k$, consistent with the governing equation. For the $v$-component, we observe that the parameter-independent term again reflects the corresponding reaction–diffusion structure present in the PDE, whereas the parameter-dependent contribution is negligible, effectively cancelling out. This indicates that the learned representation selectively attributes parameter influence where it is physically relevant, while suppressing it in components where no such dependence exists.

\section{Discussion}

In this work, we addressed the problem of learning parameterized partial differential equations in a strictly one-shot prediction setting, with particular emphasis on both in-domain and out-domain extrapolation across parameter space. Within this setting, we introduced the Late Fusion Neural Operator, a flexible framework that decouples the evolution of the state variables from the underlying system parameters. 

Across all experiments, our preposed late-fusion model has better performance than the baseline models in both in-domain and out-domain prediction settings. We hypothesize that this improvement stems from the explicit disentanglement of parameter effects and state dynamics, which may result in a more structured and potentially less nonlinear parameter dependence, thereby facilitating more robust generalization.

The interpretability analysis showed that the Late Fusion Neural Operator exhibits a degree of structural interpretability. In particular, we were able to separate parameter-dependent from parameter-independent contributions in the learned representations, which were consistent with the corresponding PDEs underlying the data. However, it is important to emphasize we did not recover explicit governing equations in a symbolic form, as the model still operates through hidden state representations and is not designed for equation discovery. A more detailed analysis of the hidden states and their relation to derivatives or PDE terms could in the future enable the use of the late-fusion framework for equation discovery.

We intentionally consider a straightforward setting without additional modelling or training components. In particular, we use a strict one-shot setup without temporal history, rely on a relatively small amount of data compared to common benchmarks \citep{Takamoto2023-my, Takamoto2022-nl}, and do not employ specialized training strategies such as noise injection \citep{Wikner2024-ka} or refinement methods \citep{Lippe2023-rq}. This design choice was made to isolate the effect of how the physical parameter is incorporated into the model, and to clearly study its influence on performance. However, this does not imply that the Late Fusion Neural Operator is limited to this setting. On the contrary, its modular structure makes it compatible with a wide range of existing neural operator architectures and training techniques. In particular, recent approaches aimed at improving long-horizon stability, such as PDE-refiner \citep{Lippe2023-rq}, as well as methods designed for irregular grid settings \citep{Lingsch2024-ry, Li2022-nq}, could be integrated into the late-fusion framework with minimal modifications, potentially extending its applicability to more challenging forecasting and extrapolation regimes.

A key takeaway from this work is that explicitly separating parameter information from state dynamics via the Late Fusion Neural Operator is a simple but effective approach that improves performance while also enabling a degree of structural interpretability. This opens up promising directions for extending this idea toward more complex extrapolation settings and more explicit forms of equation discovery.

\subsection*{Acknowledgments}
The authors acknowledge Wageningen University and Research for their investment program Data Science and Artificial Intelligence.

\section*{Data Availability Statement}
All code and data that support the findings of this study are available on Github: 
\newline
\small\href{https://github.com/evantegelen/LateFusionNeuralOperator}{https://github.com/evantegelen/LateFusionNeuralOperator}.

\bibliographystyle{plainnat}
\bibliography{references}

\appendix
\appendix
\section{Data visualisation}\label{app:visuals}
\begin{figure*}[p]
    \centering
    \includegraphics[width=1\textwidth]{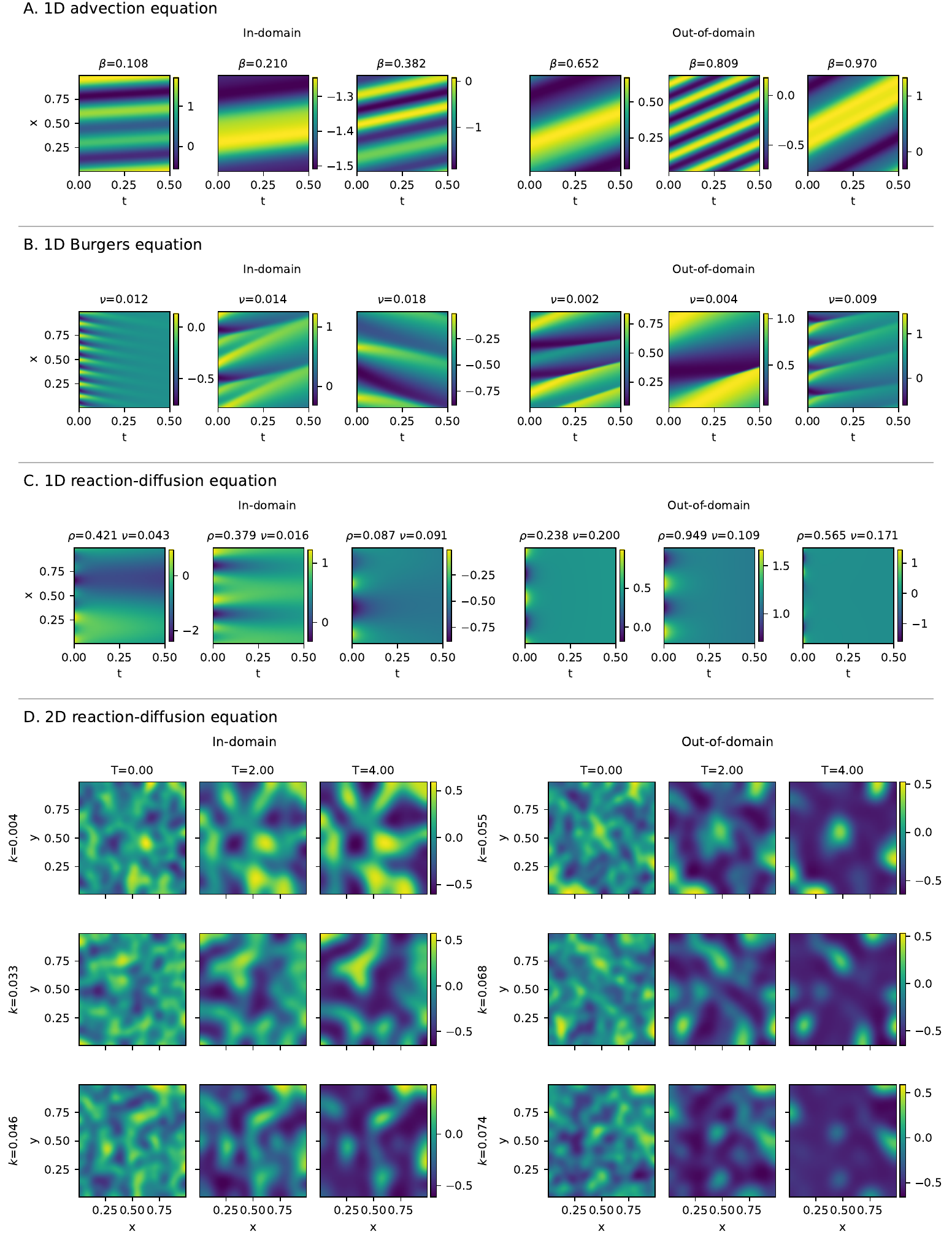}
    \caption{Visualisation of (A) 1D advection equation (B) 1D Burgers equation (C) 1D reaction-diffusion equation and (D) 2D reaction-diffusion equation. For the 1D equations trajectories corresponding to different parameter values are shown for in-domain and out-domain datasets.   
    For the 2D reaction-diffusion equation, the solutions at different time points are shown for different values of $k$.}
    \label{fig:data_visual_2d_reacdiff}
\end{figure*}

In this section, we provide visualisations of the different datasets that were used for our experiments. Figure \ref{fig:data_visual_2d_reacdiff} displays the solution trajectories for different values of the parameters for 1D advection equation, 1D Burgers equation, 1D reaction-diffusion equation and 2D reaction-diffusion equation. 

\begin{table*}[h]
\centering
\caption{Hyperparameter search space and selected values for CAPE-FNO and Late Fusion Neural Operator for each equation.}
\label{tab:hyperparameter_validation}
\vspace{0.2em}
\footnotesize
\begin{tabular}{lllcccc}
\hline
\textbf{Model} & \textbf{Hyperparameter} & \textbf{Search Space} & \textbf{advection} & \textbf{Burgers} & \textbf{Reaction-diff. (1D)} &  \textbf{Reaction-diff. (2D)}\\
\hline
CAPE-FNO 
& $\lambda_{\mathrm{CAPE}}$ 
& $\{10^{-4},10^{-6},10^{-8}\}$ 
&$10^{-4}$ &$10^{-4}$&$ 10^{-4}$&$ 10^{-4}$\\

Late Fusion 
& $\lambda_{\mathrm{Sparse}}$ 
& $\{10^{-2},10^{-3},10^{-4}\}$ 
& $10^{-4}$ &$10^{-4}$ & $10^{-4}$& $10^{-4}$\\
\hline
\end{tabular}
\end{table*}

\section{Hyperparameters and validation}
\label{app:hyperparameters}

For fair comparison, all models are trained using the same fixed training schedules and operator architectures, with additional hyperparameter tuning performed only where required.

\paragraph{Training schedules.}
For each test case all models are trained for 100 epochs using an initial learning rate of $10^{-3}$, which is reduced by a factor of two halfway. 

\paragraph{Operator architectures.}
FNO, CAPE-FNO and Late Fusion Neural Operator all rely on a neural operator backbone, and we employ the same underlying 4-level FNO architecture to ensure a fair and consistent comparison across methods. Specifically, we use:
\begin{itemize}
    \item 16 Fourier modes and width 64 for the 1D test cases,
    \item 12 Fourier modes and width 32 for the 2D test cases.
\end{itemize}  

\paragraph{Additional hyperparameter tuning.}
For the CAPE-FNO and late-fusion models, we perform additional hyperparameter selection for loss-related coefficients, with tuning carried out separately for each equation. In the case of CAPE-FNO, the weight associated with the CAPE loss term $\lambda_{\mathrm{CAPE}}$ is tuned. For the Late Fusion Neural Operator, the sparsity coefficient $\lambda_{\mathrm{sparse}}$ is selected using validation.

We reserve 10\% of the training data as a validation set for hyperparameter tuning. This validation subset is sampled uniformly across the in-domain parameter regime, ensuring that it remains representative of the training distribution. Consequently, while this strategy is suitable for selecting hyperparameters that perform well within the observed domain, it is not specifically designed to assess generalization beyond the training regime. An alternative strategy would be to construct the validation set from trajectories near the boundaries of the parameter domain, which could provide an early indication of out-of-domain generalization performance. We chose not to follow this approach, as our goal was to keep the hyperparameter selection process focused on in-domain performance rather than implicitly focusing on extrapolation performance. 

Table~\ref{tab:hyperparameter_validation} shows the search space and the values that were selected during tuning.

\section{Extra metrics}\label{appendix:extra_metrics}

For a further analysis of the Late Fusion Neural Operator Table~\ref{tab:extrametrics} provides additional metrics for the different equations. We will show the metrics that were used. All metrics were adapted from the PDE bench framework \cite{Takamoto2022-nl}. The RMSE measures the overall average deviation between predicted and true solutions:
\begin{equation}
\mathrm{RMSE}
=
\sqrt{
\frac{1}{N}
\sum_{i=1}^{N}
\left\lVert
\boldsymbol{u}_{\mathrm{pred}}^{(i)}
-
\boldsymbol{u}_{\mathrm{true}}^{(i)}
\right\rVert_2^2.
}
\end{equation}

Similarly the Boundary RMSE is the RMSE computed only taking into $u$ at boundary points. The NRMSE provides a scale-invariant version of the RMSE by normalizing with the magnitude of the true solution:
\begin{equation}
\mathrm{NRMSE}
=
\frac{\sqrt{\frac{1}{N}\sum_{i=1}^{N}\left\lVert\boldsymbol{u}_{\mathrm{pred}}^{(i)}
-\boldsymbol{u}_{\mathrm{true}}^{(i)}
\right\rVert_2^2
}
}{
\sqrt{\frac{1}{N}\sum_{i=1}^{N}\left\lVert
\boldsymbol{u}_{\mathrm{true}}^{(i)}
\right\rVert_2^2
}
+ \varepsilon.
}
\end{equation}

The Max Error captures the worst-case deviation across all samples:
\begin{equation}
\mathrm{MaxError}
=
\max_{i}
\left\lVert
\boldsymbol{u}_{\mathrm{pred}}^{(i)}
-
\boldsymbol{u}_{\mathrm{true}}^{(i)}
\right\rVert_{\infty}
\end{equation}

The Conserved Error measures how well the model preserves the spatial integral of the solution:
\begin{equation}
\mathrm{ConservedError}
=
\sqrt{
\frac{1}{N}
\left\lVert
\sum_x u_{\mathrm{pred}}(x)
-
\sum_x u_{\mathrm{true}}(x)
\right\rVert_2^2
}
\end{equation}

The Fourier RMSE evaluates the error in the spectral domain:
\begin{equation}
\mathrm{FourierRMSE}
=\sqrt{\frac{1}{N}\sum_{i=1}^{N}
\left\lVert\mathcal{F}\!\left(\boldsymbol{u}_{\mathrm{pred}}^{(i)}\right)-
\mathcal{F}\!\left(\boldsymbol{u}_{\mathrm{true}}^{(i)}\right)
\right\rVert_2^2}
\end{equation}

Table~\ref{tab:extrametrics} shows the performance of the Late-Fusion Neural Operator compared to FNO and CAPE-FNO. It can be seen that across all metrics the performance of the late-fusion approach is better than the other benchmarks.

\begin{sidewaystable*}
\caption{Metrics summary for all models and domains}
\centering
\small
\begin{tabular}{|l|ccc|ccc|}
\toprule
& \multicolumn{3}{|c|}{In-domain} & \multicolumn{3}{c|}{Out-domain} \\
& \multicolumn{3}{|c|}{} & \multicolumn{3}{c|}{} \\
Metric & FNO & CAPE-FNO & Late Fusion & FNO & CAPE-FNO & Late Fusion \\
\midrule
\multicolumn{3}{l}{\textbf{1D Advection}} \\
\midrule
Boundary & 2.62e-1 $\pm$ 2.49e-2 & 3.07e-1 $\pm$ 9.59e-2 & \textbf{3.16e-2 $\pm$ 1.30e-3} & 4.66e-1 $\pm$ 1.23e-1 & 4.46e-1 $\pm$ 2.79e-2 & \textbf{1.03e-1 $\pm$ 5.45e-3} \\
Conserved & 2.95e+1 $\pm$ 9.35e+0 & 2.90e+1 $\pm$ 9.67e+0 & \textbf{8.00e-1 $\pm$ 1.67e-1} & 3.65e+1 $\pm$ 6.70e+0 & 3.68e+1 $\pm$ 9.74e+0 & \textbf{8.54e+0 $\pm$ 3.79e+0} \\
Fourier & 5.93e+0 $\pm$ 2.01e-1 & 5.43e+0 $\pm$ 1.14e+0 & \textbf{5.38e-1 $\pm$ 2.97e-2} & 8.67e+0 $\pm$ 9.29e-1 & 8.16e+0 $\pm$ 1.53e+0 & \textbf{2.30e+0 $\pm$ 3.47e-1} \\
Max & 5.41e+0 $\pm$ 9.58e-1 & 2.87e+0 $\pm$ 6.19e-1 & \textbf{8.83e-1 $\pm$ 2.83e-2} & 5.00e+0 $\pm$ 4.67e-1 & 3.58e+0 $\pm$ 4.87e-1 & \textbf{3.54e+0 $\pm$ 3.61e-1} \\
RMSE & 4.72e-1 $\pm$ 2.34e-2 & 4.26e-1 $\pm$ 7.69e-2 & \textbf{4.75e-2 $\pm$ 2.50e-3} & 7.14e-1 $\pm$ 1.09e-1 & 6.67e-1 $\pm$ 1.21e-1 & \textbf{1.93e-1 $\pm$ 2.23e-2} \\
nRMSE & 5.73e-1 $\pm$ 2.85e-2 & 5.18e-1 $\pm$ 9.35e-2 & \textbf{5.77e-2 $\pm$ 3.04e-3} & 8.80e-1 $\pm$ 1.34e-1 & 8.22e-1 $\pm$ 1.49e-1 & \textbf{2.38e-1 $\pm$ 2.75e-2} \\
\midrule
\multicolumn{3}{l}{\textbf{1D Burgers}} \\
\midrule
Boundary & 5.41e-2 $\pm$ 2.01e-2 & 6.53e-2 $\pm$ 1.86e-2 & \textbf{2.15e-2 $\pm$ 2.06e-3} & 1.13e-1 $\pm$ 4.67e-2 & 8.55e-2 $\pm$ 2.37e-2 & \textbf{3.67e-2 $\pm$ 1.56e-3} \\
Conserved & 7.42e+0 $\pm$ 2.39e+0 & 6.53e+0 $\pm$ 1.44e+0 & \textbf{1.79e+0 $\pm$ 3.72e-1} & 1.62e+1 $\pm$ 7.12e+0 & 8.55e+0 $\pm$ 4.00e+0 & \textbf{1.63e+0 $\pm$ 2.87e-1} \\
Fourier & 1.32e+0 $\pm$ 4.08e-1 & 1.35e+0 $\pm$ 2.58e-1 & \textbf{4.67e-1 $\pm$ 2.97e-2} & 2.51e+0 $\pm$ 8.93e-1 & 1.85e+0 $\pm$ 5.80e-1 & \textbf{7.85e-1 $\pm$ 9.62e-3} \\
Max & 1.77e+0 $\pm$ 1.40e-1 & 1.96e+0 $\pm$ 3.53e-1 & \textbf{1.30e+0 $\pm$ 1.85e-1} & 2.46e+0 $\pm$ 3.96e-1 & 2.46e+0 $\pm$ 2.70e-1 & \textbf{1.95e+0 $\pm$ 1.53e-1} \\
RMSE & 1.02e-1 $\pm$ 3.18e-2 & 1.09e-1 $\pm$ 2.08e-2 & \textbf{3.91e-2 $\pm$ 2.02e-3} & 1.83e-1 $\pm$ 5.95e-2 & 1.50e-1 $\pm$ 4.28e-2 & \textbf{6.87e-2 $\pm$ 7.40e-4} \\
nRMSE & 1.55e-1 $\pm$ 4.85e-2 & 1.66e-1 $\pm$ 3.17e-2 & \textbf{5.96e-2 $\pm$ 3.08e-3} & 2.60e-1 $\pm$ 8.45e-2 & 2.13e-1 $\pm$ 6.07e-2 & \textbf{9.75e-2 $\pm$ 1.05e-3} \\
\midrule
\multicolumn{3}{l}{\textbf{1D Reaction-diffusion}} \\
\midrule
Boundary & 1.41e-1 $\pm$ 1.33e-2 & 1.57e-1 $\pm$ 1.24e-2 & \textbf{2.71e-2 $\pm$ 1.91e-3} & 2.37e-1 $\pm$ 8.60e-2 & 1.54e-1 $\pm$ 3.60e-2 & \textbf{2.84e-2 $\pm$ 1.03e-2} \\
Conserved & 2.72e+1 $\pm$ 6.39e+0 & 2.97e+1 $\pm$ 3.24e+0 & \textbf{6.11e+0 $\pm$ 1.37e+0} & 4.26e+1 $\pm$ 1.86e+1 & 3.43e+1 $\pm$ 7.55e+0 & \textbf{1.11e+1 $\pm$ 9.12e+0} \\
Fourier & 3.92e+0 $\pm$ 6.29e-1 & 4.27e+0 $\pm$ 2.96e-1 & \textbf{8.87e-1 $\pm$ 1.70e-1} & 5.76e+0 $\pm$ 2.16e+0 & 4.68e+0 $\pm$ 8.24e-1 & \textbf{1.40e+0 $\pm$ 1.13e+0} \\
Max & 1.05e+1 $\pm$ 2.21e+0 & 9.61e+0 $\pm$ 8.30e-1 & \textbf{3.84e+0 $\pm$ 1.34e+0} & 6.29e+0 $\pm$ 1.79e+0 & 5.53e+0 $\pm$ 1.14e+0 & \textbf{1.54e+0 $\pm$ 8.20e-1} \\
RMSE & 2.77e-1 $\pm$ 3.27e-2 & 3.01e-1 $\pm$ 1.49e-2 & \textbf{6.29e-2 $\pm$ 1.10e-2} & 3.90e-1 $\pm$ 1.30e-1 & 3.19e-1 $\pm$ 5.01e-2 & \textbf{8.98e-2 $\pm$ 7.10e-2} \\
nRMSE & 3.53e-1 $\pm$ 4.18e-2 & 3.83e-1 $\pm$ 1.90e-2 & \textbf{8.02e-2 $\pm$ 1.41e-2} & 5.00e-1 $\pm$ 1.66e-1 & 4.09e-1 $\pm$ 6.43e-2 & \textbf{1.15e-1 $\pm$ 9.12e-2} \\
\midrule
\multicolumn{3}{l}{\textbf{2D Reaction-diffusion}} \\
\midrule
Boundary & 2.93e-2 $\pm$ 1.79e-3 & 4.87e-2 $\pm$ 4.70e-3 & \textbf{1.92e-2 $\pm$ 1.27e-3} & 5.43e-2 $\pm$ 3.69e-2 & 7.62e-2 $\pm$ 2.57e-3 & \textbf{1.91e-2 $\pm$ 4.04e-4} \\
Conserved & 1.01e+0 $\pm$ 3.03e-2 & 3.30e+0 $\pm$ 1.55e-1 & \textbf{2.49e-1 $\pm$ 8.38e-3} & 7.06e+0 $\pm$ 9.93e+0 & 5.78e+0 $\pm$ 6.62e-1 & \textbf{3.02e-1 $\pm$ 1.06e-2} \\
Fourier & 2.55e-1 $\pm$ 6.03e-3 & 6.41e-1 $\pm$ 3.10e-2 & \textbf{8.65e-2 $\pm$ 4.67e-3} & 1.28e+0 $\pm$ 1.71e+0 & 1.05e+0 $\pm$ 1.04e-1 & \textbf{8.66e-2 $\pm$ 3.39e-3} \\
Max & 6.43e-1 $\pm$ 9.11e-2 & 8.01e-1 $\pm$ 6.81e-2 & \textbf{4.51e-1 $\pm$ 9.95e-2} & 1.32e+0 $\pm$ 1.16e+0 & 8.56e-1 $\pm$ 6.26e-2 & \textbf{4.86e-1 $\pm$ 1.17e-1} \\
RMSE & 2.83e-2 $\pm$ 8.23e-4 & 6.29e-2 $\pm$ 3.51e-3 & \textbf{1.03e-2 $\pm$ 5.86e-4} & 1.19e-1 $\pm$ 1.52e-1 & 9.87e-2 $\pm$ 8.43e-3 & \textbf{9.92e-3 $\pm$ 4.89e-4} \\
nRMSE & 1.17e-1 $\pm$ 3.41e-3 & 2.61e-1 $\pm$ 1.46e-2 & \textbf{4.25e-2 $\pm$ 2.43e-3} & 3.62e-1 $\pm$ 4.63e-1 & 2.99e-1 $\pm$ 2.56e-2 & \textbf{3.01e-2 $\pm$ 1.48e-3} \\
\bottomrule
\end{tabular}\label{tab:extrametrics}
\end{sidewaystable*}

\end{document}